\documentclass{article}

\usepackage{arxiv}
\usepackage[utf8]{inputenc}
\usepackage[T1]{fontenc}
\usepackage{microtype}
\usepackage{amsmath,amssymb,mathtools}
\usepackage{booktabs}
\usepackage{tabularx}
\usepackage{longtable}
\usepackage{array}
\usepackage{graphicx}
\usepackage{natbib}
\usepackage{hyperref}
\usepackage{xcolor}
\usepackage{enumitem}
\usepackage[nohyperlinks]{acronym}
\usepackage{tikz}
\usetikzlibrary{arrows.meta,positioning,fit}

\newcommand{\E}{\mathbb{E}}
\newcommand{\LCAS}{\operatorname{CAS}_{L}}
\newcommand{\SLCAS}{\widetilde{\operatorname{CAS}}_{L}}
\newcommand{\GICAS}{\operatorname{CAS}_{G,\mathrm{inst}}}
\newcommand{\GMCAS}{\operatorname{CAS}_{G,\mathrm{mass}}}
\newcommand{\FCAS}{\operatorname{F\!CAS}}

\newacro{CAS}{Causal Attribution Score}
\newacro{XAI}{explainable artificial intelligence}
\newacro{SHAP}{SHapley Additive exPlanations}
\newacro{LIME}{Local Interpretable Model-agnostic Explanations}
\newacro{DML}{double/debiased machine learning}
\newacro{AIPW}{augmented inverse-probability weighting}
\newacro{ATE}{average treatment effect}
\newacro{CATE}{conditional average treatment effect}
\newacro{MAE}{mean absolute error}
\newacro{IQR}{interquartile range}
\newacro{MAD}{median absolute deviation}

\title{CAS: A Causal Attribution Score for Local and Global Explainable Artificial Intelligence}

\author{
Michael Georgiades\\
Department of Computer Science\\
Neapolis University Pafos\\
Cyprus
\and
Charalambia Varnava\\
CaSToRC\\
The Cyprus Institute\\
Nicosia, Cyprus
}

\renewcommand{\shorttitle}{CAS: A Causal Attribution Score}

\hypersetup{
  colorlinks=true,
  linkcolor=blue,
  citecolor=blue,
  urlcolor=blue,
  pdftitle={CAS: A Causal Attribution Score for Local and Global Explainable Artificial Intelligence},
  pdfauthor={Michael Georgiades, Charalambia Varnava}
}

\begin{document}
\maketitle

\begin{abstract}
Predictive explanation methods attribute a model output; they do not, by themselves, attribute an intervention effect on the real-world outcome. We introduce the \ac{CAS}, a compact score architecture for causal explanation. CAS starts from an identified interventional coalition game, allocates the joint intervention contrast with causal Shapley contributions and converts those raw outcome-scale effects into Local CAS, Signed Local CAS and two complementary Global CAS summaries. The innovation is not a new Shapley formula, but a local-to-global causal reporting layer with an explicit intervention target.

In the known-truth benchmark, eight repeated primary-interaction simulations ($n=2{,}200$ each, three actions) gave mean Local CAS \ac{MAE} $0.107$ for coalition-aware CAS, compared with $0.173$ for one-at-a-time normalisation and $0.213$ for a global normalised absolute \ac{ATE} vector. The paired advantage over one-at-a-time normalisation increased from $-0.003$ under additivity to $0.091$ under strong interactions. On both empirical DoubleML datasets, 401(k) eligibility/net financial assets ($n=9{,}915$) and Pennsylvania reemployment bonus/unemployment duration ($n=5{,}099$), predictive \ac{SHAP}/TreeSHAP rankings differed materially from Feature-CAS rankings of treatment-effect modifiers. In Pennsylvania, \texttt{dep1} (exactly one dependent) moved from predictive global rank 13 to Feature-CAS rank 2 and was the leading local Feature-CAS modifier. These results isolate the added value of separating \emph{what predicts the outcome} from \emph{what explains heterogeneity in an estimated causal effect}.
\end{abstract}


\section{Motivation and contribution}\label{sec:motivation}

\ac{LIME} and SHAP explain why a fitted model produces a prediction \citep{ribeiro2016should,lundberg2017unified}. In both methods, the target is predictive:
\begin{equation}
f(x)-f_{\mathrm{base}}.
\end{equation}
An intervention-oriented explanation has a different target:
\begin{equation}
\E\!\left[Y(a_1)-Y(a_0)\mid X=x\right].
\end{equation}
Here, $x$ is the covariate profile at which the model is explained, $f_{\mathrm{base}}$ is a reference prediction (e.g.\ the model's average output) and $Y(a_1)$ and $Y(a_0)$ are potential outcomes under target and baseline intervention profiles $a_1=a(Q)$ and $a_0=a(\varnothing)$, formalised in Section~\ref{sec:definitions}. A feature can be highly predictive because it is a proxy, confounder, descendant or marker of treatment selection without being an actionable cause. Conversely, an intervention can have a real effect, while receiving little predictive importance.

\begin{figure}[t]
\centering
\begin{tikzpicture}[
  node distance=6mm and 5mm,
  every node/.style={font=\small,align=center},
  box/.style={draw,rounded corners,minimum height=10mm,minimum width=28mm,fill=gray!6},
  arr/.style={-{Latex[length=2mm]},thick}
]
\node[box] (spec) {Causal specification\\actions $Q$, baseline/target};
\node[box,right=of spec] (game) {Interventional game\\$v_x(S)$};
\node[box,right=of game] (phi) {Raw causal allocation\\$\phi_j^C(x)$};
\node[box,right=of phi] (local) {Local CAS\\magnitude + sign};
\node[box,below=of local] (global) {Global CAS\\instance / effect-mass};
\draw[arr] (spec) -- (game);
\draw[arr] (game) -- (phi);
\draw[arr] (phi) -- (local);
\draw[arr] (local) -- (global);
\node[draw,dashed,rounded corners,fit=(spec)(game)(phi)(local)(global),inner sep=3mm,
      label=below:{\footnotesize causal estimand $\rightarrow$ allocation $\rightarrow$ communication}] {};
\end{tikzpicture}
\caption{CAS is a score architecture built on an identified interventional target. The raw causal allocation remains in outcome units; Local and Global CAS provide relative explanatory summaries.}
\label{fig:casflow}
\end{figure}
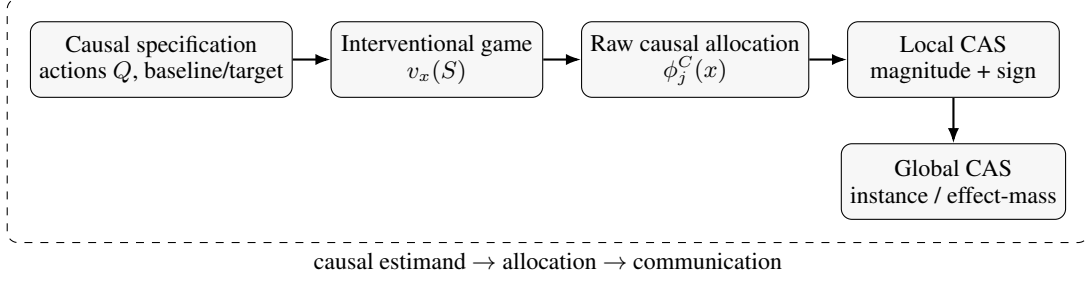

Prior causal Shapley work incorporates causal structure into Shapley feature attribution \citep{heskes2020causal}. In this work, we propose CAS. CAS does \emph{not} claim a new Shapley value. Its novelty is the score layer built around a declared intervention game:
\begin{enumerate}[leftmargin=*,nosep]
\item define the causal quantity being explained;
\item allocate its joint intervention contrast in outcome units;
\item convert the allocation into comparable local scores;
\item aggregate local scores into distinct global summaries.
\end{enumerate}
This separation is the central contribution. The score is meaningful only under a declared intervention set and standard causal identification assumptions: well-defined interventions, consistency, conditional exchangeability, positivity/joint support and an interference specification \citep{pearl2001direct}. Figure~\ref{fig:casflow} summarises the resulting progression from causal specification to local and global attribution.

\section{CAS definitions}\label{sec:definitions}

Let $Q=\{1,\ldots,q\}$ denote the declared actionable concepts. Each action $j$ has a baseline level $a_j^0$ and a target level $a_j^1$. For coalition $S\subseteq Q$, let $a(S)$ activate the target level for actions in $S$ and the baseline level otherwise. For profile $x$, define
\begin{equation}
v_x(S)=\E[Y(a(S))\mid X=x]
\label{eq:value}
\end{equation}
and the joint intervention contrast
\begin{equation}
\Delta_Q(x)=v_x(Q)-v_x(\varnothing).
\label{eq:joint}
\end{equation}

\subsection{Raw causal contribution}

CAS uses the classical Shapley value \citep{shapley1953value} applied to the \emph{interventional} game:
\begin{equation}
\phi_j^C(x)=
\sum_{S\subseteq Q\setminus\{j\}}
\frac{|S|!(q-|S|-1)!}{q!}
\left[v_x(S\cup\{j\})-v_x(S)\right].
\label{eq:phi}
\end{equation}
The raw contributions are in outcome units and satisfy
\begin{equation}
\sum_{j\in Q}\phi_j^C(x)=\Delta_Q(x).
\label{eq:efficiency}
\end{equation}
Equation~\eqref{eq:phi} is an allocation mechanism; CAS is the score layer defined next.

\subsection{Local CAS}

Let
\begin{equation}
A_+(x)=\sum_{k\in Q}|\phi_k^C(x)|
\end{equation}
be local causal mass. When $A_+(x)=0$, the joint intervention effect is exactly zero at $x$ ($\Delta_Q(x)=0$ by Equation~\eqref{eq:efficiency}). Local CAS is undefined at such instances and they are excluded from the population averages in Equations~\eqref{eq:ginst}--\eqref{eq:gmass}. For $A_+(x)>0$,
\begin{equation}
\boxed{
\LCAS^{(j)}(x)=
\frac{|\phi_j^C(x)|}{\sum_k|\phi_k^C(x)|}
}
\label{eq:lcas}
\end{equation}
is the fraction of absolute causal mass assigned to action $j$. It is non-negative and sums to one. Direction is retained by
\begin{equation}
\boxed{
\SLCAS^{(j)}(x)=
\frac{\phi_j^C(x)}{\sum_k|\phi_k^C(x)|}
}
\label{eq:slcas}
\end{equation}
with
\begin{equation}
\sum_j\SLCAS^{(j)}(x)=\frac{\Delta_Q(x)}{A_+(x)}\in[-1,1],
\end{equation}
which follows from the triangle inequality,
$|\Delta_Q(x)|=|\sum_j\phi_j^C(x)|\leq\sum_j|\phi_j^C(x)|=A_+(x)$. The signed sum therefore exposes reinforcement or cancellation rather than behaving as a probability.

\subsection{Global CAS}

Local and global explanation answer different questions. The first Global CAS gives every profile equal weight:
\begin{equation}
\boxed{
\GICAS^{(j)}=\E_X[\LCAS^{(j)}(X)]
}
\label{eq:ginst}
\end{equation}
and answers: \emph{which action is typically prominent across individuals?}

The second weights by causal-effect mass:
\begin{equation}
\boxed{
\GMCAS^{(j)}=
\frac{\E_X[|\phi_j^C(X)|]}
{\sum_k\E_X[|\phi_k^C(X)|]}
}
\label{eq:gmass}
\end{equation}
and answers: \emph{which action accounts for the greatest absolute causal effect in the population?}

This distinction is necessary because an action may be modest but frequently important, or rare but very large. Both global scores sum to one, but they represent different population summaries.

\subsection{Why coalition-aware CAS is more than normalised \ac{CATE}}

If the intervention game is additive,
\begin{equation}
v_x(S)=v_x(\varnothing)+\sum_{j\in S}\tau_j(x),
\end{equation}
then the marginal contribution of action $j$ is independent of coalition context and
\begin{equation}
\phi_j^C(x)=\tau_j(x),\qquad
\LCAS^{(j)}(x)=\frac{|\tau_j(x)|}{\sum_k|\tau_k(x)|}.
\label{eq:additive}
\end{equation}
Thus CAS deliberately reduces to normalised conditional treatment effects when there are no interactions. Its coalition-aware added value appears when the effect of one intervention depends on which other interventions are active.

\section{Estimation and Feature-CAS}\label{sec:estimation}

For the empirical illustrations we use two datasets distributed with the Python \texttt{DoubleML} package \citep{bach2022doubleml}. The 401(k) sample contains $n=9{,}915$ observations from the 1991 Survey of Income and Program Participation and follows the 401(k) treatment-response application of \citet{abadie2003semiparametric}; we use eligibility (\texttt{e401}) as treatment and net financial assets (\texttt{net\_tfa}) as outcome. The Pennsylvania sample contains $n=5{,}099$ observations from the Reemployment Bonus experiment analysed by \citet{bilias2000sequential}; we use the treatment-group indicator (\texttt{tg}) and log unemployment duration (\texttt{inuidur1}). The exact variable definitions used in the analysis are listed in Appendix~\ref{app:variables}; no optional polynomial feature expansion is used.

Treatment effects are estimated with cross-fitting and an orthogonal \ac{AIPW} score, following \ac{DML} \citep{chernozhukov2018double}. Cross-fitted AIPW estimation relies on the identification assumptions stated in Section~\ref{sec:motivation}: consistency, conditional exchangeability given $X$ and positivity/overlap in treatment assignment. Writing $\widehat\mu_{di}=\widehat\mu_d(X_i)$ and $\widehat e_i=\widehat e(X_i)$, the binary-treatment score is
\begin{equation}
\widehat\psi_i=\widehat\mu_{1i}-\widehat\mu_{0i}+\frac{D_i(Y_i-\widehat\mu_{1i})}{\widehat e_i}-\frac{(1-D_i)(Y_i-\widehat\mu_{0i})}{1-\widehat e_i}.
\label{eq:aipw}
\end{equation}
The ATE is $\widehat\tau_{\mathrm{ATE}}=n^{-1}\sum_i\widehat\psi_i$. A second cross-fitted learner estimates the heterogeneous effect
\begin{equation}
\widehat\tau_i=g^{(-k(i))}(X_i),
\end{equation}
where observation $i$ is predicted by a model trained without its fold.

For feature-level comparison with predictive \ac{XAI}, we use \emph{Feature-CAS}, which explains the \emph{estimated treatment-effect surface}, not the outcome prediction:
\begin{equation}
\widehat\tau_i=b_i+\sum_{j=1}^{p}\gamma_{ij}.
\label{eq:featurecas}
\end{equation}
Here, $p$ is the number of pre-treatment covariates and $b_i$ is the fold-specific TreeSHAP base value determined from the held-out fold's CATE learner and its training-fold background. Because the second-stage CATE learner is a Random Forest, $\gamma_{ij}$ is computed with TreeSHAP \citep{lundberg2017unified} on the exact fold-specific learner that generated $\widehat\tau_i$. The implementation checks additivity and reconstructs the stored cross-fitted CATE to numerical tolerance. It does not explicitly enumerate all $2^p$ covariate subsets.

The corresponding local and global feature scores are
\begin{equation}
\FCAS_{L,j}(x_i)=\frac{|\gamma_{ij}|}{\sum_k|\gamma_{ik}|},\qquad
\FCAS_{G,j}=\frac{\E|\gamma_j(X)|}{\sum_k\E|\gamma_k(X)|}.
\end{equation}
These are \emph{effect-modifier attributions}: they identify which pre-treatment covariates explain heterogeneity in the estimated causal effect. They do not claim that directly manipulating a covariate such as age produces $\gamma_{ij}$.

Feature-CAS is deliberately not a direct instance of the interventional game $v_x(S)$ in Equation~\eqref{eq:value}: pre-treatment covariates are effect modifiers, not declared actions with baseline/target levels $a_j^0,a_j^1$. Equation~\eqref{eq:featurecas} applies the same Shapley-type allocation logic to a different object, the already-estimated scalar $\widehat\tau_i$. Predictive SHAP/TreeSHAP and LIME are evaluated on a Random-Forest outcome model and explain $f(x)$; Feature-CAS explains $\widehat\tau(x)$. The comparison is therefore intentionally between different explanation targets.

\paragraph{Fair local-profile selection.}
To avoid selecting a profile because one explainer makes it extreme, all local comparisons use the same observed row chosen from $X$ alone. Let $m_j=\operatorname{median}_i X_{ij}$ and let $s_j=\operatorname{IQR}_i X_{ij}$ (with \ac{MAD}/standard-deviation fallbacks if the \ac{IQR} is zero). We select
\begin{equation}
i_{\mathrm{typ}}=\arg\min_i\left\{\frac{1}{p}\sum_{j=1}^{p}\left(\frac{X_{ij}-m_j}{s_j}\right)^2\right\}^{1/2}.
\label{eq:typical}
\end{equation}
No outcome, treatment, model prediction, SHAP/LIME value, ATE, CATE or Feature-CAS quantity enters this selection rule. The same row is then explained by all three local methods.

\paragraph{Exact enumeration versus exact reconstruction.}
The core known-truth CAS benchmark below has $q=3$ declared interventions and therefore exactly enumerates all $2^q=8$ intervention coalitions before applying Equation~\eqref{eq:phi}. Empirical Feature-CAS instead uses the tree-specific TreeSHAP algorithm on $p=9$ or $p=15$ effect modifiers. Accordingly, ``exact Feature-CAS reconstruction'' means exact additive reconstruction of the fitted fold-specific CATE-model output up to numerical tolerance; it does not mean exhaustive $2^p$ enumeration, nor does it replace the causal identification assumptions for $\widehat\tau$.

\section{Results}\label{sec:results}

\subsection{Known-truth validation of the CAS core}

The strongest test of the core score is a structural benchmark in which every interventional coalition value is known (Appendix~\ref{app:dgp}). Table~\ref{tab:truth} reports repeated-seed results for the primary interaction scenario. Raw contribution MAE evaluates $\phi_j^C$ directly; Local CAS MAE evaluates the normalised score; top-action accuracy is
\begin{equation}
\Pr\!\left[\arg\max_j|\widehat\phi_j^C(x)|=\arg\max_j|\phi_j^{C,\star}(x)|\right].
\end{equation}
Each row in Table~\ref{tab:truth} is the mean over $R=8$ independent seeds with $n=2{,}200$ instances per seed. Brackets are descriptive 95\% Monte Carlo intervals across seeds.

\begin{table*}[t]
\caption{Repeated known-truth primary-interaction benchmark ($R=8$, $n=2{,}200$, $q=3$, eight exactly enumerated intervention coalitions per profile). Lower is better for MAE; higher is better for top-action accuracy. Values are mean [95\% Monte Carlo interval].}
\label{tab:truth}
\centering
\small
\begin{tabular}{lccc}
\toprule
Method & Raw $\phi$ \acs{MAE} & Local CAS \acs{MAE} & Top-action accuracy\\
\midrule
Coalition-aware CAS & \textbf{0.321 [0.312, 0.330]} & \textbf{0.107 [0.100, 0.114]} & \textbf{0.803 [0.787, 0.819]}\\
One-at-a-time normalisation & 0.500 [0.480, 0.520] & 0.173 [0.165, 0.182] & 0.775 [0.752, 0.798]\\
Global normalised $|ATE|$ & 0.617 [0.606, 0.627] & 0.213 [0.196, 0.230] & 0.762 [0.751, 0.773]\\
\bottomrule
\end{tabular}
\end{table*}

\begin{figure}[t]
\centering
\includegraphics[width=0.82\linewidth]{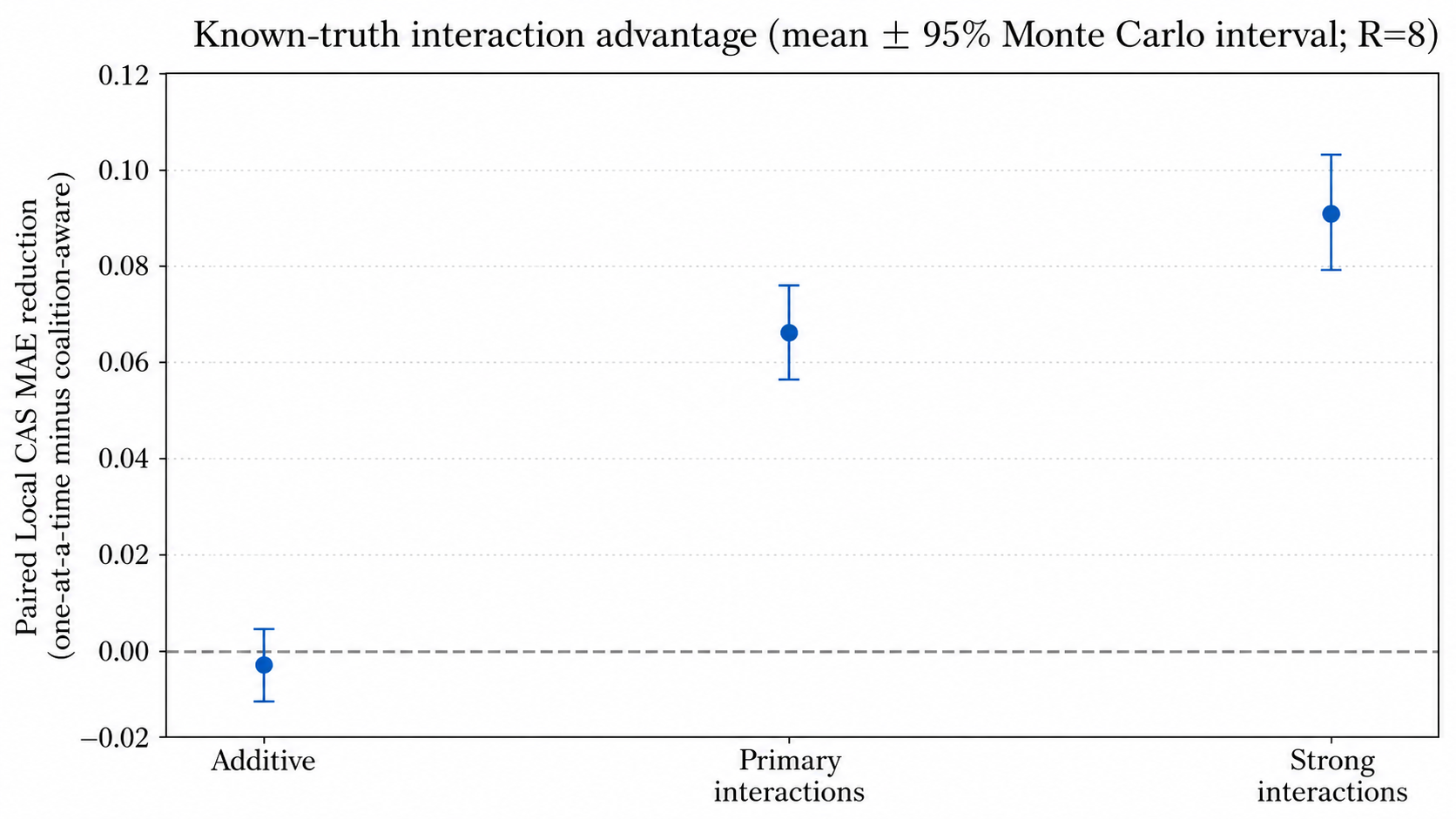}
\caption{Paired Local CAS MAE advantage over one-at-a-time normalisation across additive, primary-interaction and strong-interaction known-truth scenarios. Points are mean paired reductions over $R=8$ seeds; bars are descriptive 95\% Monte Carlo intervals.}
\label{fig:interactionadvantage}
\end{figure}

The interaction-strength stress test is a \emph{separate repeated experiment}, not the subtraction of two entries in Table~\ref{tab:truth}. The paired Local CAS MAE reduction (one-at-a-time minus coalition-aware CAS) was $0.066$ [0.057, 0.075] in the primary-interaction scenario and $0.091$ [0.080, 0.103] under strong interactions. With interactions removed it was $-0.003$ [$-0.010$, 0.003], statistically indistinguishable from zero at this Monte Carlo resolution (Figure~\ref{fig:interactionadvantage}). This directly matches Equation~\eqref{eq:additive}: coalition context adds information when interactions matter and should not create an artificial advantage in an additive game.

\subsection{Real-data comparison with predictive explanations}

The population figures use 301 explained profiles per dataset, while the underlying samples contain $9{,}915$ 401(k) observations and $5{,}099$ Pennsylvania observations. Table~\ref{tab:ranks} summarises the global estimand change and adds conditional row-bootstrap intervals ($B=500$) for the displayed population feature-mass scores. The numerical values in Table~\ref{tab:ranks} are bootstrap means, whereas the feature ordering in the trajectory plots is determined from the fitted-sample global masses; the small numerical differences do not change the reported ranks. These intervals quantify finite-profile variability conditional on the fitted models; they do not propagate refitting uncertainty. TreeSHAP gives the same top-three predictive ordering as KernelSHAP in both datasets and nearly identical mass estimates.

\begin{table*}[t]
\caption{Global feature-mass comparison on the 301 displayed profiles. Values are mean mass [95\% conditional row-bootstrap interval], $B=500$. The rank shift is KernelSHAP global rank $\rightarrow$ Feature-CAS global rank.}
\label{tab:ranks}
\centering
\scriptsize
\begin{tabularx}{\textwidth}{l X X l}
\toprule
Dataset & Predictive SHAP top 3 & Feature-CAS top 3 & Illustrative rank shift\\
\midrule
401(k) & \texttt{inc} 0.315 [0.288,0.342]; \texttt{pira} 0.302 [0.290,0.313]; \texttt{age} 0.138 [0.125,0.149]
& \texttt{inc} 0.395 [0.365,0.426]; \texttt{age} 0.157 [0.141,0.173]; \texttt{educ} 0.104 [0.090,0.117]
& \texttt{pira}: 2 $\rightarrow$ 7\\
Pennsylvania bonus & \texttt{agelt35} 0.207 [0.199,0.214]; \texttt{black} 0.157 [0.144,0.170]; \texttt{agegt54} 0.131 [0.121,0.141]
& \texttt{female} 0.123 [0.112,0.134]; \texttt{dep1} 0.121 [0.106,0.137]; \texttt{lusd} 0.108 [0.096,0.118]
& \texttt{dep1}: 13 $\rightarrow$ 2\\
\bottomrule
\end{tabularx}
\end{table*}

\begin{figure*}[t]
\centering
\includegraphics[width=\textwidth]{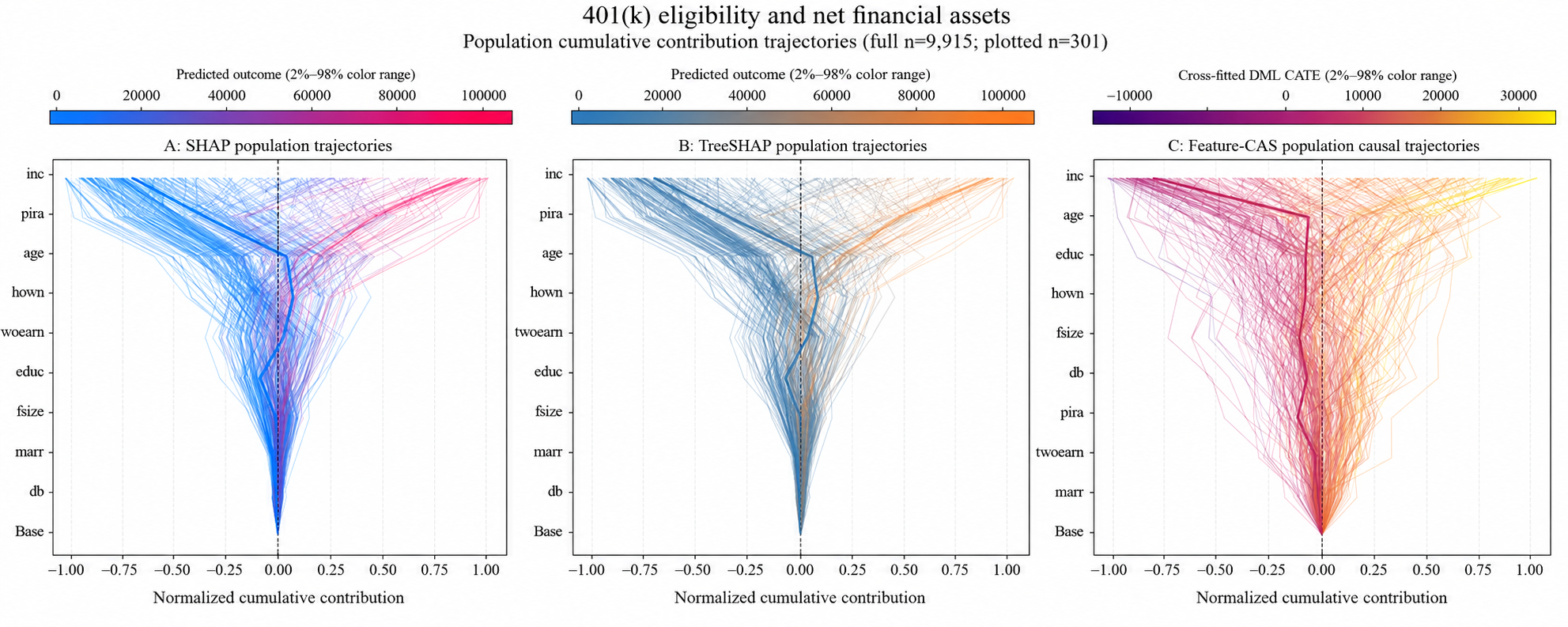}
\caption{401(k): population cumulative contribution trajectories (full $n=9{,}915$; 301 profiles displayed). SHAP and TreeSHAP explain the Random-Forest outcome model; Feature-CAS explains the exact cross-fitted DML-CATE surface.}
\label{fig:401kpopulation}
\end{figure*}

\begin{figure*}[t]
\centering
\includegraphics[width=\textwidth]{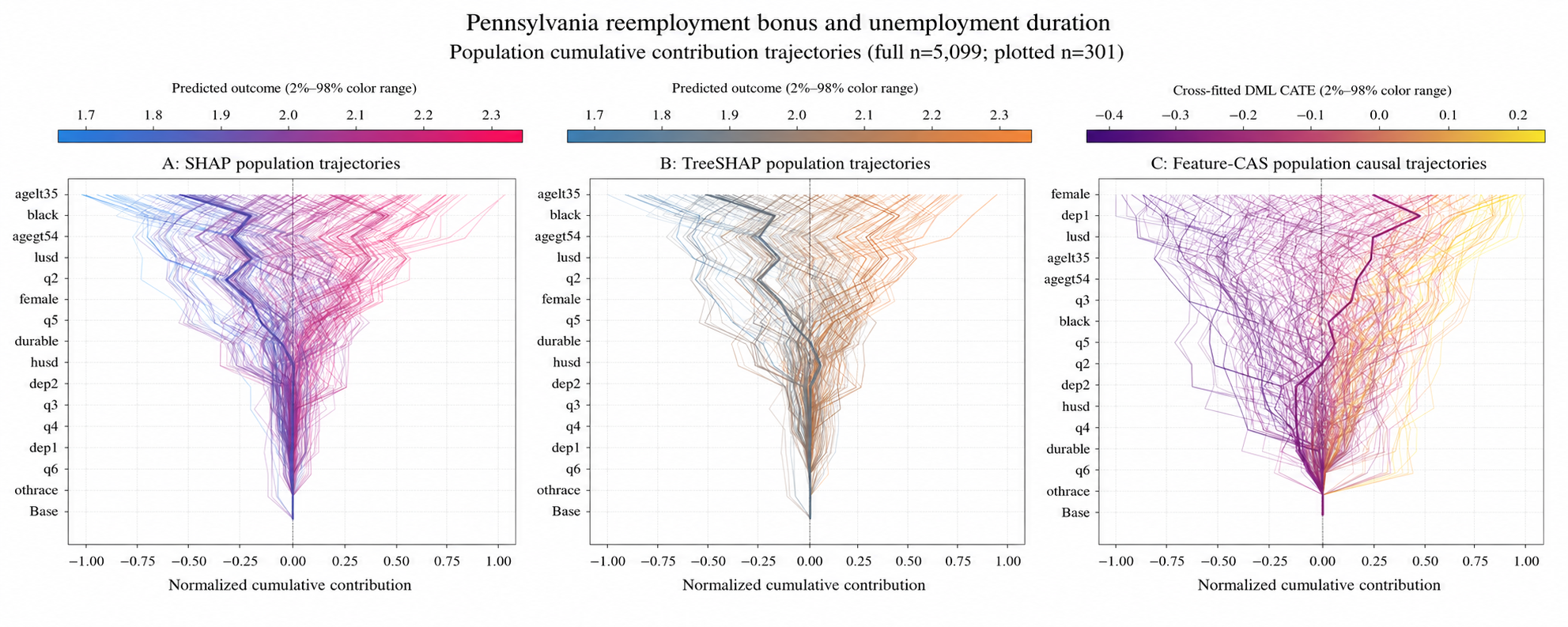}
\caption{Pennsylvania bonus: population cumulative contribution trajectories (full $n=5{,}099$; 301 profiles displayed). Predictive SHAP/TreeSHAP and Feature-CAS produce different feature orderings because they explain different estimands.}
\label{fig:bonuspopulation}
\end{figure*}

Figures~\ref{fig:401kpopulation} and~\ref{fig:bonuspopulation} show the full population-trajectory comparison with enlarged manuscript-scale axis labels. In 401(k), SHAP and TreeSHAP emphasise \texttt{inc}, \texttt{pira} and \texttt{age}, whereas Feature-CAS shifts mass toward \texttt{inc}, \texttt{age} and \texttt{educ}. In Pennsylvania, predictive explanations emphasise \texttt{agelt35}, \texttt{black} and \texttt{agegt54}, whereas Feature-CAS places \texttt{female}, \texttt{dep1} and \texttt{lusd} at the top of the causal-effect-modifier ordering.

\begin{figure*}[t]
\centering
\includegraphics[width=\textwidth]{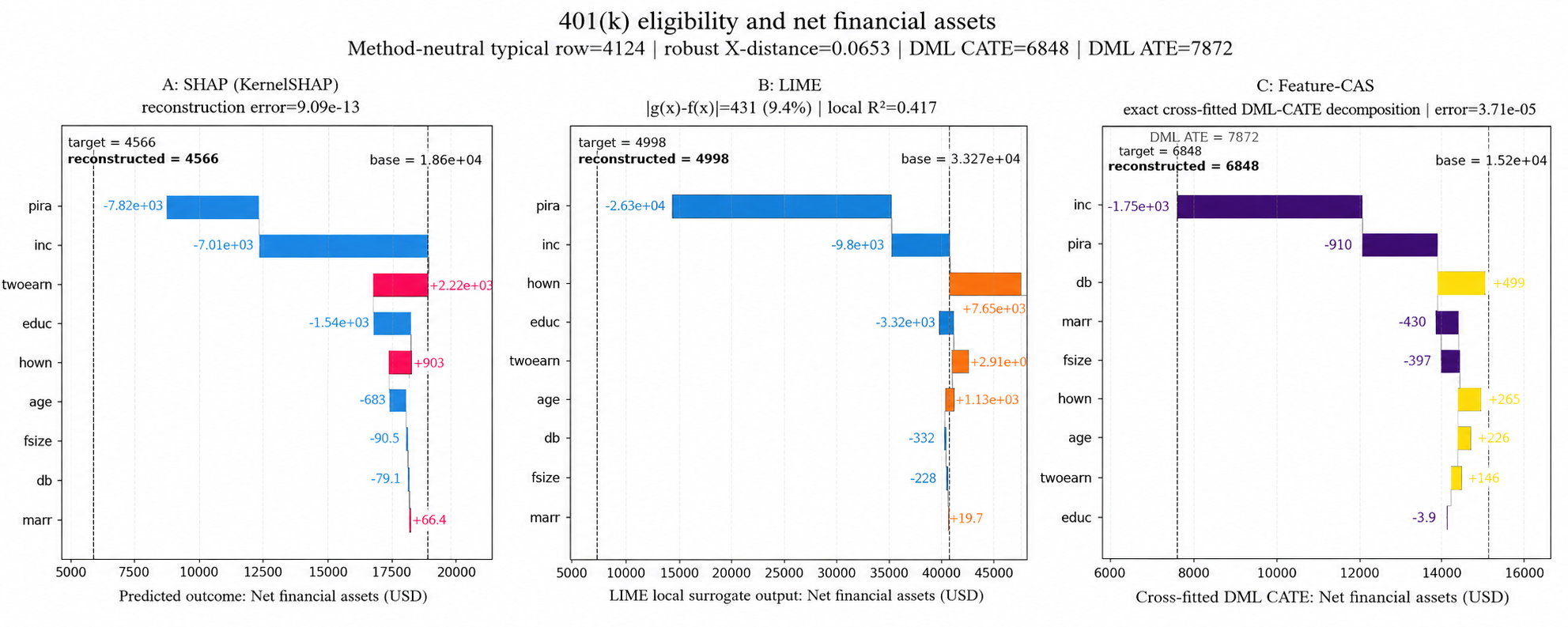}
\caption{401(k) method-neutral typical profile. SHAP and LIME explain the predictive outcome model; Feature-CAS explains the exact fold-specific cross-fitted DML CATE. All three methods explain the same row selected from $X$ alone.}
\label{fig:401klocal}
\end{figure*}

The local comparison now uses the method-neutral typical profile in Equation~\eqref{eq:typical}, which removes the earlier asymmetry of selecting a profile for an extreme CATE. For 401(k), the selected row is 4124 (robust $X$-distance 0.0653): the reference-run DML CATE is approximately \$6,848 and the ATE \$7,872. SHAP and LIME both rank \texttt{pira} and \texttt{inc} first, whereas Feature-CAS ranks \texttt{inc}, \texttt{pira} and \texttt{db}. Feature-CAS reconstructs the stored cross-fitted CATE with absolute error $3.71\times10^{-5}$ outcome units, corresponding to relative error $5.4\times10^{-9}$. LIME's local surrogate differs from the Random-Forest prediction by 9.4\%, with local weighted $R^2=0.417$ (Figure~\ref{fig:401klocal}). Because this row is selected as the most central observed profile under the robust $X$-distance rule, the remaining LIME discrepancy should be interpreted as local surrogate fidelity rather than as deliberate selection of a causal outlier.

\begin{figure*}[t]
\centering
\includegraphics[width=\textwidth]{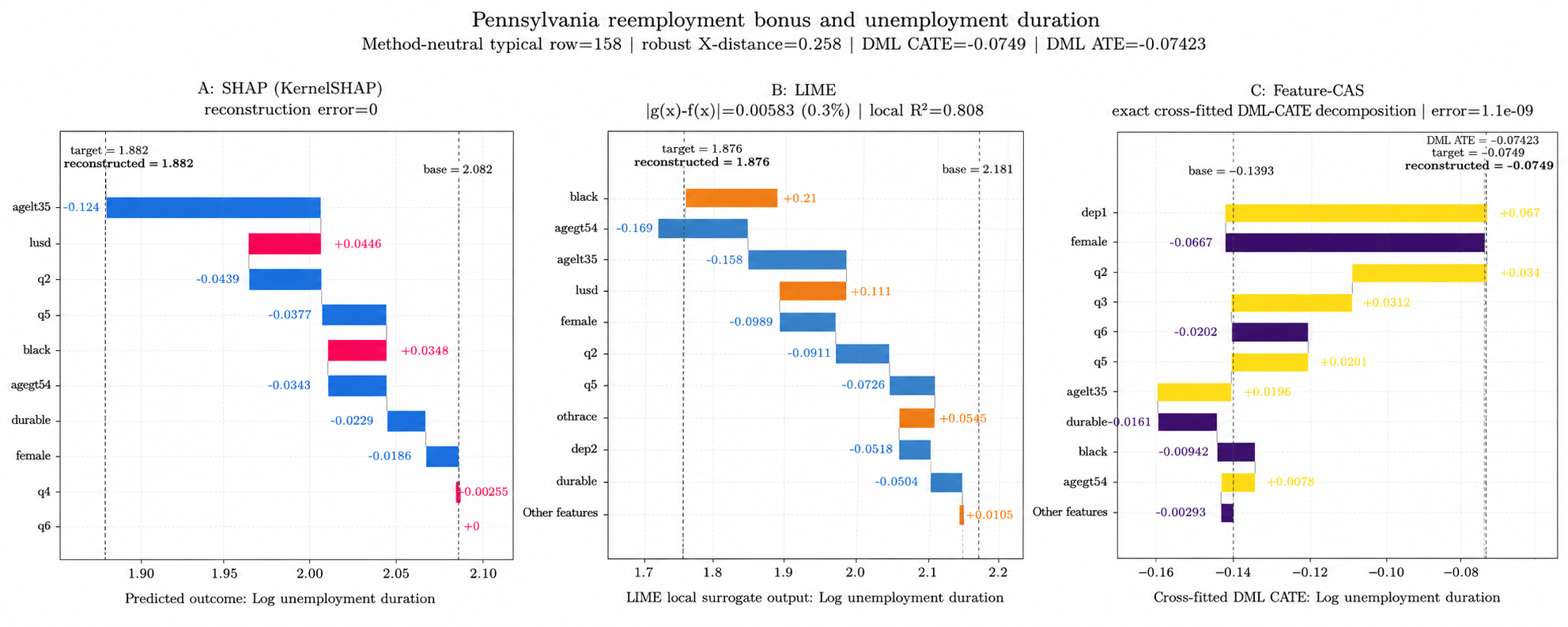}
\caption{Pennsylvania method-neutral typical profile. The leading Feature-CAS modifiers are \texttt{dep1}, \texttt{female} and \texttt{q2}, whereas SHAP and LIME emphasise different predictive features.}
\label{fig:bonuslocal}
\end{figure*}

For Pennsylvania, the selected row is 158 (robust $X$-distance 0.258): the reference-run DML CATE is $-0.0749$ and the ATE $-0.07423$. Predictive SHAP ranks \texttt{agelt35}, \texttt{lusd} and \texttt{q2} highest; LIME ranks \texttt{black}, \texttt{agegt54} and \texttt{agelt35}; Feature-CAS ranks \texttt{dep1}, \texttt{female} and \texttt{q2}. Feature-CAS reconstruction error is $1.10\times10^{-9}$ outcome units, corresponding to relative error $1.46\times10^{-8}$. LIME is locally close to the Random-Forest prediction, with a 0.31\% surrogate discrepancy and local $R^2=0.808$ (Figure~\ref{fig:bonuslocal}). The contrast with 401(k) illustrates that LIME fidelity is profile- and surface-dependent and should be reported rather than assumed.

\paragraph{Split-seed stability.}
To give a separate view of model/split variability, the DML pipeline was rerun over five independent split/model seeds, while holding the same method-neutral row fixed (Table~\ref{tab:stability}). The ATE and CATE values printed in Figures~\ref{fig:401klocal} and~\ref{fig:bonuslocal} are the single reference-run estimates used to construct those exact decompositions; Table~\ref{tab:stability} instead reports means across five reruns, so the table values are not expected to equal the figure annotations exactly. These are descriptive algorithmic-stability intervals, not formal causal confidence intervals.

\begin{table}[t]
\caption{DML split-seed stability over $R=5$ repetitions. Values are mean $\pm$ standard deviation [descriptive 95\% Monte Carlo interval].}
\label{tab:stability}
\centering
\scriptsize
\begin{tabular}{lcc}
\toprule
Dataset & ATE & Typical-row CATE\\
\midrule
401(k) & 7,915 $\pm$ 130 [7,753, 8,077] & 7,218 $\pm$ 1,821 [4,957, 9,479]\\
Pennsylvania & $-0.07427\pm0.00296$ [$-0.07795,-0.07059$] & $-0.1548\pm0.1198$ [$-0.3036,-0.0061$]\\
\bottomrule
\end{tabular}
\end{table}

\section{What CAS adds}

The results support three narrow claims.

\paragraph{1. CAS changes the explained object.}
Predictive SHAP/LIME answer ``what contributed to this prediction?''. CAS answers ``how is an identified intervention effect allocated?''. Feature-CAS correspondingly asks which pre-treatment variables explain heterogeneity in that causal effect.

\paragraph{2. Local and global causal importance are not interchangeable.}
Local CAS is a compositional explanation for one profile. Instance-Balanced Global CAS describes typical local prominence, whereas Effect-Mass Global CAS describes population causal mass. One cannot in general replace these quantities with a single global $|ATE|$ ranking.

\paragraph{3. Coalition awareness matters exactly when interactions matter.}
Under additivity, CAS reduces to normalised CATEs. Under interaction, one-at-a-time effects omit coalition context. The repeated known-truth benchmark and interaction stress test show that the recovery advantage grows with interaction strength and vanishes under additivity. This is the principal evidence that coalition-aware CAS is more than a relabelling of one-at-a-time treatment effects.

CAS remains an explanatory score, not an optimisation objective. Raw signed effects and the outcome direction must be retained when an action is to be chosen.

\section{Conclusion}

CAS provides a minimal local-to-global language for causal attribution. It begins with a declared interventional target, preserves a raw outcome-scale causal allocation, then separates magnitude, sign, individual prominence and population effect mass. The framework deliberately collapses to normalised conditional treatment effects in additive settings and departs from them only when coalition context changes marginal effects.

The evidence supports this narrow claim from two directions. In known truth, coalition-aware recovery improves as interaction strength increases and shows no artificial advantage under additivity. In the two empirical datasets, method-neutral local profiles and population summaries show that predictive SHAP/LIME importance can differ substantially from the covariates that explain heterogeneity in an estimated DML treatment effect. Together, these results establish the added value of CAS without treating predictive attribution and causal attribution as interchangeable tasks.

\appendix
\section{Dataset variables}\label{app:variables}

Table~\ref{tab:variables} lists the exact treatment, outcome and base covariates used in the empirical analysis. The 401(k) data source is the 1991 Survey of Income and Program Participation application used by \citet{abadie2003semiparametric}, while the Pennsylvania Reemployment Bonus experiment is documented by \citet{bilias2000sequential}. Variable names and preprocessing follow the DoubleML data interface \citep{bach2022doubleml}.

\renewcommand{\arraystretch}{1.08}
\begin{longtable}{p{0.16\textwidth} p{0.14\textwidth} p{0.62\textwidth}}
\caption{Variables used in the empirical analysis.}\label{tab:variables}\\
\toprule
Dataset / role & Variable & Definition\\
\midrule
\endfirsthead
\toprule
Dataset / role & Variable & Definition\\
\midrule
\endhead
401(k), outcome & \texttt{net\_tfa} & Net financial assets, US dollars.\\
401(k), treatment & \texttt{e401} & Eligibility for a 401(k) plan.\\
401(k), covariate & \texttt{age} & Age.\\
401(k), covariate & \texttt{inc} & Income, US dollars.\\
401(k), covariate & \texttt{educ} & Education in years.\\
401(k), covariate & \texttt{fsize} & Family size.\\
401(k), covariate & \texttt{marr} & Married indicator.\\
401(k), covariate & \texttt{twoearn} & Two-earner household indicator.\\
401(k), covariate & \texttt{db} & Defined-benefit pension indicator.\\
401(k), covariate & \texttt{pira} & Individual retirement account participation indicator.\\
401(k), covariate & \texttt{hown} & Home-owner indicator.\\
\midrule
Pennsylvania, outcome & \texttt{inuidur1} & DoubleML-preprocessed log duration of the first unemployment spell; the raw measure is in weeks.\\
Pennsylvania, treatment & \texttt{tg} & Binary treatment-group indicator in the preprocessed bonus-analysis sample.\\
Pennsylvania, covariate & \texttt{female} & Female indicator.\\
Pennsylvania, covariate & \texttt{black} & Black-race indicator.\\
Pennsylvania, covariate & \texttt{othrace} & Other-race indicator: non-white, non-black, non-Hispanic.\\
Pennsylvania, covariate & \texttt{dep1} & Indicator for exactly one dependent.\\
Pennsylvania, covariate & \texttt{dep2} & Indicator for exactly two dependents.\\
Pennsylvania, covariate & \texttt{q2} & Enrollment-quarter 2 indicator; quarter 1 is the reference.\\
Pennsylvania, covariate & \texttt{q3} & Enrollment-quarter 3 indicator; quarter 1 is the reference.\\
Pennsylvania, covariate & \texttt{q4} & Enrollment-quarter 4 indicator; quarter 1 is the reference.\\
Pennsylvania, covariate & \texttt{q5} & Enrollment-quarter 5 indicator; quarter 1 is the reference.\\
Pennsylvania, covariate & \texttt{q6} & Enrollment-quarter 6 indicator; quarter 1 is the reference.\\
Pennsylvania, covariate & \texttt{agelt35} & Indicator for claimant age below 35.\\
Pennsylvania, covariate & \texttt{agegt54} & Indicator for claimant age above 54.\\
Pennsylvania, covariate & \texttt{durable} & Indicator for an occupation in durable manufacturing.\\
Pennsylvania, covariate & \texttt{lusd} & Indicator for filing at a low-unemployment / short-duration site (Coatesville, Reading or Lancaster).\\
Pennsylvania, covariate & \texttt{husd} & Indicator for filing at a high-unemployment / short-duration site (Lewistown, Pittston or Scranton).\\
\bottomrule
\end{longtable}

\section{Known-truth structural benchmark}\label{app:dgp}

The reproducible known-truth benchmark uses six independent standard-normal covariates $X_1,\ldots,X_6$, three binary interventions $A_1,A_2,A_3$ and
\begin{align}
\mu(X) &= 0.6X_1-0.35X_2+0.25X_3^2+0.20\sin X_4+0.15X_5X_6,\\
\tau_1(X) &= 1+0.65X_1-0.20X_2,\\
\tau_2(X) &= 0.35+0.70\sin X_2+0.25X_3,\\
\tau_3(X) &= 0.55-0.75\mathbb{1}(X_1<0)+0.35X_4.
\end{align}
Interaction functions are
\begin{align}
\gamma_{12}(X)&=s\{0.90+0.30\mathbb{1}(X_1>0)\},\\
\gamma_{23}(X)&=s(-0.75+0.25X_2),\\
\gamma_{13}(X)&=0.45s\tanh(X_3+X_5),
\end{align}
where $s\in\{0,1,1.8\}$ gives the additive, primary-interaction and strong-interaction scenarios. The structural outcome is
\begin{equation}
Y=\mu(X)+\sum_{j=1}^{3}A_j\tau_j(X)+\gamma_{12}A_1A_2+\gamma_{23}A_2A_3+\gamma_{13}A_1A_3+\varepsilon,
\qquad \varepsilon\sim N(0,0.8^2).
\end{equation}
The three treatment propensities are
\begin{align}
e_1(X)&=\operatorname{logit}^{-1}(-0.15+0.60X_1-0.25X_4),\\
e_2(X)&=\operatorname{logit}^{-1}(0.10-0.45X_2+0.30X_5),\\
e_3(X)&=\operatorname{logit}^{-1}(-0.05+0.40X_3-0.25X_6),
\end{align}
with $A_j\sim\operatorname{Bernoulli}(e_j(X))$ conditionally on $X$. For each profile, the benchmark evaluates all eight intervention coalitions exactly, computes the true causal Shapley allocation by Equation~\eqref{eq:phi} and compares it with allocations obtained from an estimated Random-Forest outcome surface. The results in Section~\ref{sec:results} use $n=2{,}200$ and $R=8$ independent seeds per scenario.

\bibliographystyle{plainnat}
\bibliography{references}

@inproceedings{ribeiro2016should,
  title={{``Why Should I Trust You?'': Explaining the Predictions of Any Classifier}},
  author={Ribeiro, Marco Tulio and Singh, Sameer and Guestrin, Carlos},
  booktitle={Proceedings of the 22nd ACM SIGKDD International Conference on Knowledge Discovery and Data Mining},
  pages={1135--1144},
  year={2016},
  doi={10.1145/2939672.2939778}
}

@inproceedings{lundberg2017unified,
  title={A Unified Approach to Interpreting Model Predictions},
  author={Lundberg, Scott M. and Lee, Su-In},
  booktitle={Advances in Neural Information Processing Systems},
  volume={30},
  pages={4765--4774},
  year={2017}
}

@incollection{shapley1953value,
  title={A Value for $n$-Person Games},
  author={Shapley, Lloyd S.},
  booktitle={Contributions to the Theory of Games II},
  editor={Kuhn, Harold W. and Tucker, Albert W.},
  pages={307--317},
  publisher={Princeton University Press},
  year={1953}
}

@inproceedings{heskes2020causal,
  title={Causal Shapley Values: Exploiting Causal Knowledge to Explain Individual Predictions of Complex Models},
  author={Heskes, Tom and Sijben, Evi and Bucur, Ioan Gabriel and Claassen, Tom},
  booktitle={Advances in Neural Information Processing Systems},
  volume={33},
  pages={4778--4789},
  year={2020}
}

@article{chernozhukov2018double,
  title={Double/Debiased Machine Learning for Treatment and Structural Parameters},
  author={Chernozhukov, Victor and Chetverikov, Denis and Demirer, Mert and Duflo, Esther and Hansen, Christian and Newey, Whitney and Robins, James},
  journal={The Econometrics Journal},
  volume={21},
  number={1},
  pages={C1--C68},
  year={2018},
  doi={10.1111/ectj.12097}
}

@article{bach2022doubleml,
  title={{DoubleML} -- An Object-Oriented Implementation of Double Machine Learning in Python},
  author={Bach, Philipp and Chernozhukov, Victor and Kurz, Malte S. and Spindler, Martin},
  journal={Journal of Machine Learning Research},
  volume={23},
  number={53},
  pages={1--6},
  year={2022}
}

@inproceedings{pearl2001direct,
  title={Direct and Indirect Effects},
  author={Pearl, Judea},
  booktitle={Proceedings of the Seventeenth Conference on Uncertainty in Artificial Intelligence},
  pages={411--420},
  year={2001},
  publisher={Morgan Kaufmann}
}

@article{abadie2003semiparametric,
  title={Semiparametric Instrumental Variable Estimation of Treatment Response Models},
  author={Abadie, Alberto},
  journal={Journal of Econometrics},
  volume={113},
  number={2},
  pages={231--263},
  year={2003},
  doi={10.1016/S0304-4076(02)00201-4}
}

@article{bilias2000sequential,
  title={Sequential Testing of Duration Data: The Case of the Pennsylvania `Reemployment Bonus' Experiment},
  author={Bilias, Yannis},
  journal={Journal of Applied Econometrics},
  volume={15},
  number={6},
  pages={575--594},
  year={2000},
  doi={10.1002/jae.579}
}

\end{document}